\documentclass{article}

\usepackage[final]{neurips_2024}

\usepackage[utf8]{inputenc} 
\usepackage[T1]{fontenc}    
\usepackage{hyperref}       
\usepackage{url}            
\usepackage{booktabs}       
\usepackage{amsfonts}       
\usepackage{nicefrac}       
\usepackage{microtype}      
\usepackage{xcolor}         
\usepackage{graphicx}
\usepackage{CJK}
\usepackage{multirow}
\usepackage{tabularx}
\usepackage{lineno}
\usepackage{listings}
\usepackage{color}

\title{MooER: LLM-based Speech Recognition and Translation Models from Moore Threads}

\author{%
  Zhenlin Liang$^{1*}$ \quad Junhao Xu$^2$\thanks{Equal Contribution} \quad Yi Liu \\
  \textbf{Yichao Hu} \quad \textbf{Jian Li} \quad \textbf{Yajun Zheng} \quad \textbf{Meng Cai}\thanks{Corresponding Author} \quad \textbf{Hua Wang}\\
  Moore Threads\\
  \texttt{\{zhenlin.liang, junhao.xu, yi.liu\}@mthreads.com} \\
  \texttt{\{yichao.hu, jian.li, yajun.zheng, meng.cai, hua.wang\}@mthreads.com}
}

\begin{document}

\maketitle

\begin{abstract}
In this paper, we present MooER, a LLM-based large-scale automatic speech recognition (ASR) / automatic speech translation (AST) model of Moore Threads. A 5000h pseudo labeled dataset containing open source and self collected speech data is used for training. We achieve performance comparable to other open source models trained with up to hundreds of thousands of hours of labeled speech data. Meanwhile, experiments conducted on Covost2 Zh2en testset [1] suggest that our model outperforms other open source Speech LLMs. A BLEU score of 25.2 can be obtained. The main contributions of this paper are summarized as follows. First, this paper presents a training strategy for encoders and LLMs on speech related tasks (including ASR and AST) using a small size of pseudo labeled data without any extra manual annotation and selection. Second, we release our ASR and AST models and plan to open-source our training code and strategy in the near future. Moreover, a model trained on 8wh scale training data is planned to be released later on. 
\end{abstract}

\section{Motivation}
In May 2024, OpenAI released GPT-4o, which supports end-to-end speech inputs and outputs, 
 pioneering the end-to-end speech interation technology based on LLMs. From then on, speech, the most natural human-machine interaction modality, has entered a new age of 'the GPT Moment'. To this end, researchers have been continuously exploring large scale speech models. However, there are still uncertainties in the model structure (such as the selection of LLMs, the selection of speech encoders, and the connection relationship between speech encoders and LLMs), the training methods (such as which stages of training are divided, which parameters are adjusted separately, the required data size, computing resources, and costs), and so on. Especially in the open source community, existing work related to speech large models can be mainly divided into two categories: the first one uses open source training datasets to validate their performance on academic benchmarks, such as Salmonn [2]; The second type uses a large amount of data and training resources to train models for multiple speech-related tasks, such as whisper [3], SeamlessM4T [4], Qwen-audio [5], SenseVoice [6], SpeechLlama [7], etc. At present, there are few large-scale speech models that can realize industrial-scale applications for specific vertical classes under the constraints of resources.

Our released work applies the speech large model technology to the tasks of speech ASR and AST in the following two aspects. Firstly, in terms of model structure and training, we used the open-source Qwen2-7B-instruct model [8] and open-source Paraformer [9] model encoder for model initialization. During the training procedure, only the speech adapter and LLM Lora [10] parameters are fine tuned. Moreover, to improve the training speed, stability, as well as inference speed, optimization techniques such as DeepSpeed [11], Dataloader acceleration, gradient checkpoint, gradient accumulation, and BF16 acceleration are also applied. Secondly, as for the training resources, we used only 8 domestically produced S4000 GPUs developed by Moore Threads for computation. Our large audio understanding model was trained using 5000 hours (composed of a combination of partially open-source datasets and pseudo-labeled datasets) within 38 hours. Thirdly, a CER of 4.21\% on 6 Mandarin test sets and a WER of 17.98\% on 6 English test sets were obtained on ASR tasks. Meanwhile, we alse achieved a BLEU score of 25.2 on the Covost2 Zh2en translation test set, which meets the requirement of industrial application. Fourthly, the training and inference of this work are based on Moore Threads GPUs. As far as we know, this is the first speech large-scale model to use domestic GPUs for training and inference. We also demonstrated the effectiveness of an audio comprehension model trained on 80000 hours of data, which achieved a CER of 3.5\% on 6 Mandarin test sets and a WER of 12.66\% on 6 English test sets for ASR tasks.

We release the inference code and the model trained on 5000 hours of data in this work, and plan to open source training code and a model trained on 80000 hours of data in the near future. We hope this work can contribute to the community in terms of the evolution of methods and technical implementation of speech modeling.

\section{Method}

\begin{figure}[htbp]
    \centering
    \includegraphics[width=1.0\linewidth]{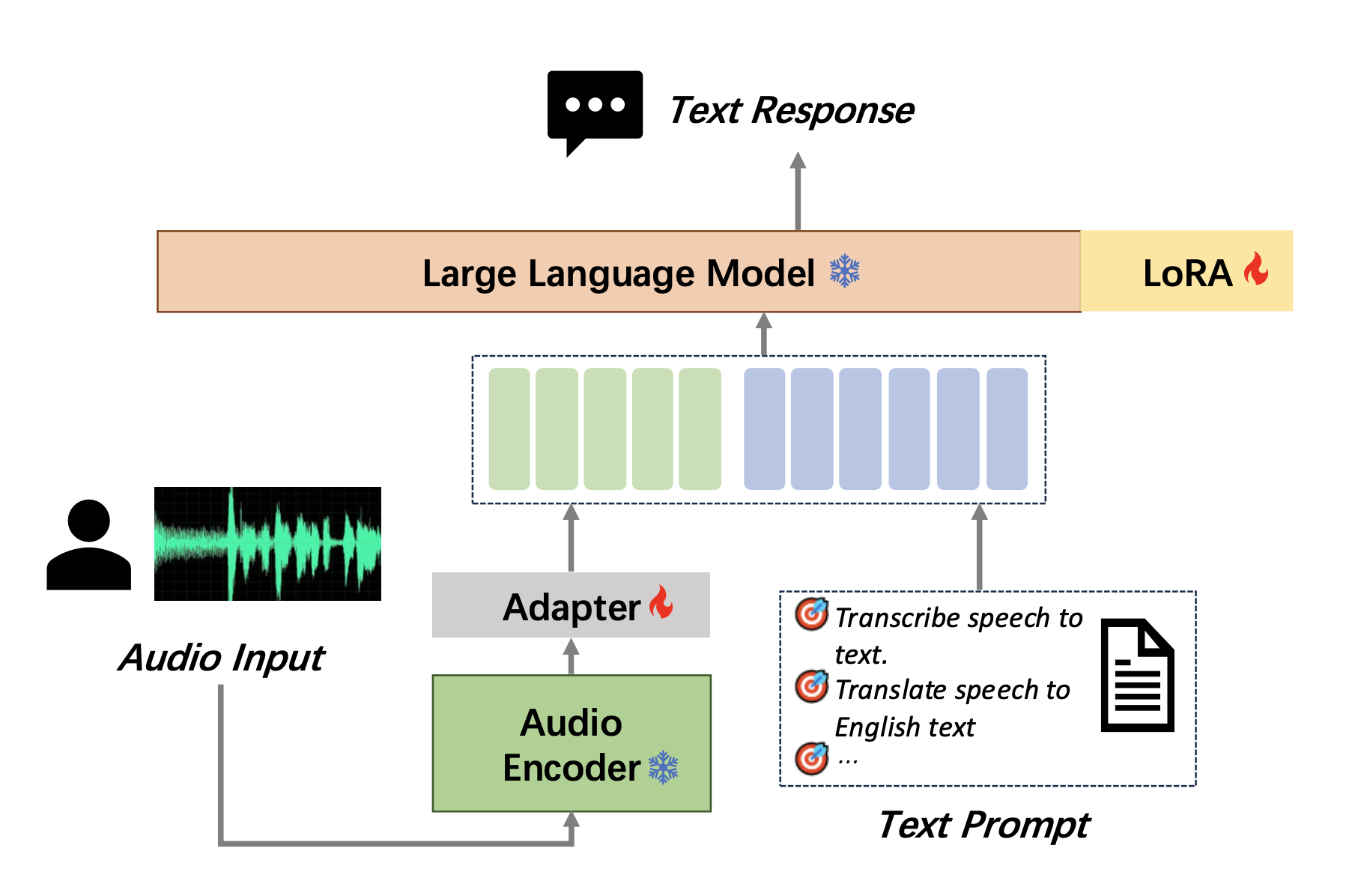}
    \caption{Model Structure}
    \label{fig:model}
\end{figure}

MooER has been greatly inspired by the following amazing works and teams: SLAM-LLM [12], we thank all the contributors for open-sourcing. We concentrate more on the tasks of ASR and AST, and have put forward some corespondibg optimization methods for the model structure, as well as the training strategy on large-scale industrial data. For example, DeepSpeed, Dataloader acceleration, gradient checkpoint, gradient accumulation, BF16 training, are combined during the fine-tuning procedure under 5000 hours of training data. The proposed model consists of encoder, adapter, and decoder (LLM) as in figure\ref{fig:model}. Encoder implements feature extraction and embedding on audio, adapter performs down-sampling of audio modality and fusion of text modality. LLM performs corresponding tasks based on the input audio and text prompts, such as ASR, AST, etc.

We have tried different encoders, including Whisper, W2v-Bert2.0 [13], Paraformer, etc. In the end, we chose Paraformer's encoder to model the audio. We use Qwen2-7B-instruct as the LLM Decoder. The audio will be downsampled using LFR(Low Frame Rate) with a downsample rate of 6 before entering the Paraformer. The audio vector output by the Encoder will be further down sampled by the Adapter, thereby reducing the density of fusion with text embedding. The adapter will downsample the audio by 2 and pass it through two linear layers to obtain the final audio prompt embedding. So, the modeling granularity of audio is $120ms$ per embedding. After splicing the audio prompt embedding and the text prompt embedding, they are sent to LLM for corresponding recognition or translation operations. During the training process, the Encoder always has fixed parameters, while the Adapter and LLM (Lora) participate in training and gradient updates.

\section{Dataset}
We constructed the MT5K training dataset containing a total of 5000 hours speech data from the following source in table \ref{tab:mt5k}:

\begin{table}[htbp]
  \caption{MT5K dataset}
  \label{tab:mt5k}
  \centering
  \begin{tabular}{lc}
    \toprule
    \textbf{source}   & \textbf{duration/$h$} \\
    \midrule
    Aishell2 [14]  & 137    \\
    Librispeech [15]    & 131      \\
    multi\_cn [16]  & 100  \\
    wenetspeech [17]   & 1361 \\
    in-house        &   3274  \\
    \bottomrule
  \end{tabular}
\end{table}

The data in the open-source dataset is randomly selected from the entire dataset. In house data is collected internally, and its ASR pseudo labels are obtained through the recording file recognition interface of a third-party cloud service. We will call the third-party translation interface with the corresponding ASR pseudo label to obtain the corresponding AST pseudo label. We did not use any data filtering methods to select crawled data and pseudo labels, which may potentially reduce manual processes in industrial production environments. We also present the performance of our ASR model trained on 80000 hours of internal data on the test set.

\section{Experiments}

Our model structure is as follows. We use a Paraformer large encoder as our audio encoder. Paraformer has only 158M parameters, which is lighter and has no padding to 30s limit (whisper), making both training and inference more efficient. The output of the audio encoder will be further down sampled by the adapter, thereby reducing the density of fusion with text embedding. The adapter will downsample the audio by 2 and pass it through two linear layers to obtain the final audio prompt embedding. Our audio feature frames are shifted by 10ms, and the front-end uses LFR for downsampling with a downsampling rate of 6. So, the modeling granularity of audio is 120ms per embedding. The audio embedding will be directly spliced before the text prompt embedding and sent to LLM for training. We use Qwen2-7B instruction as our LLM Decoder.

The final input form of LLM is: 

\textit{<\textbf{Speech Embedding}><|im\_start|>system \textbackslash n You\ are\ a helpful assistant.}

\textit{<|im\_end|>\textbackslash n<|im\_start|>user\textbackslash nTranslate speech to english text.<|im\_end|>\textbackslash n<|im\_start|>assistant\textbackslash n
\begin{CJK}{UTF8}{gbsn}
\textit{你叫什么名字？}
\end{CJK}\textbackslash n{What's your name?}<|im\_end|>}

During the training process, we will freeze the parameters of the encoder and train the adapter and LLM (lora). Our experiment found that if LLM is involved in the training process, its semantic understanding ability can be utilized to improve the final audio understanding effect. The configuration for training Lora is as in Appendix A. Ultimately, 2\% of LLM parameters will be involved in training.
Our model has the following parameter scale as in table \ref{tab:params}.

\begin{table}[htbp]
  \caption{Model Params}
  \label{tab:params}
  \centering
  \begin{tabular}{lcc}
    \toprule
    \textbf{Module}   & \textbf{Params($M$)} & \textbf{trainable Params($M$)} \\
    \midrule
    Encoder  & 158 & 0    \\
    Adapter    &  9.44 & 9.44    \\
    LLM  & 7615.62 & 161.48 \\
    \bottomrule
  \end{tabular}
\end{table}

To enhance training speed and reduce GPU memory usage, we train using the Zero2 optimization within the DeepSpeed framework. We have observed that the Paraformer tends to experience upward overflow when using Fp16, therefore we utilize Bf16 for both training and inference. We used 8 domestically produced S4000 GPUs developed by Moore Threads and trained them based on the KUAE framework, which took 38 hours. Our Deepspeed configuration is shown in Applendix B.

\subsection{ASR}

We compared the ASR performance on 6 Mandarin test sets and 6 English test sets using Paraformer-large, SenseVoice small, Qwen-audio, WhiperV3, and SeamlessM4T2. We also present the performance of our ASR model trained on 80000 hours of internal data on the test set. We compared the data used by different models during the ASR/AST stages. The ASR results are shown in table ~\ref{tab:asr}.

\begin{itemize}
    \item \textit{Paraformer-large}: 60,000 hours ASR data
    \item \textit{SenseVoice small}: 300,000 hours ASR data
    \item \textit{Qwen-audio}: 53,000 hours ASR data + 3700 hours S2TT data + ...
    \item \textit{WhisperV3}: 1000,000 hours weakly labels, 4000,000 hours pseudo labels
    \item \textit{SeamlessM4T2}: 351,000 hours S2TT data, 145,000 hours S2ST data
    \item \textit{MooER-5K}: 5,000 hours pseudo labels
    \item \textit{MooER-80K}: 80,000 hours pseudo labels
\end{itemize}
\begin{table}[htbp]
  \label{tab:asr}
  \centering
  \caption{ASR results. \textit{aishell2: aishell2\_ios; t\_mdata: test\_magicdata; f\_c: fleurs\_cmn; lib\_t: librispeech\_test; f\_e: fleurs\_eng; giga: gigaspeech}}
  \begin{tabularx}{\textwidth}{l|c|ccccccc}
    \toprule
       & testsets & {Paraform.} & {SenseV.} & {Qwen} & {Whisper} & {Seamless} & {MooER} & {MooER} \\
       & & large & small & audio & v3 & M4T2 & 5k & 8w \\

    \midrule
    \multirow{7}{*}{ZH} &  {aishell1} & 1.93 & 3.03 & 1.43 & 7.86 &  4.09 & 1.93 & 1.25   \\ 
    & {aishell2}  & 2.85 & 3.79 & 3.57 & 5.38 & 4.81 & 3.17 & 2.67 \\ 
    & thchs & {3.99} &  {5.17} & {4.86} & {9.06} & {7.14} & {4.11} & {3.14} \\
    & {t\_mdata} & {3.66} & {3.81} & {5.31} & {8.36} & {9.69} & {3.48} & {2.52} \\
    & f\_c\_dev & 5.56 & 6.39 & 10.54 & 4.54 & 7.12 & 5.81 & 5.23 \\
    & f\_c\_test & 6.92 &  7.36 &  11.07 & 5.24 &7.66 & 6.77 & 6.18 \\ 
    & \textbf{AVG.} & \textbf{4.15} & \textbf{4.93} & \textbf{6.13} & \textbf{6.74} & \textbf{6.75} & \textbf{4.21} & \textbf{3.50} \\
    \midrule
    \multirow{7}{*}{EN} & lib\_t\_clean & 14.15 & 4.07 & 2.15 & 3.42 & 2.77 & 7.78 & 4.11 \\
    & lib\_t\_other & 22.99 & 8.26 & 4.68 & 5.62 & 5.25 & 15.25 & 9.99 \\
    & f\_e\_dev & 24.93 & 12.92 & 22.53 & 11.63 & 11.36 & 18.89 & 13.32 \\
    & f\_e\_test & 26.81 & 13.41 & 22.51 & 12.57 & 11.82 & 20.41 & 14.97 \\
    & giga\_dev & 24.23 & 19.44 & 12.96 & 19.18 & 28.01 & 23.46 & 16.92 \\
    & giga\_test & 23.07 & 16.65 & 13.26 & 22.34 & 28.65 & 22.09 & 16.64 \\
    & \textbf{AVG.} & \textbf{22.7} & \textbf{12.46} & \textbf{13.02} & \textbf{12.46} & \textbf{14.64} & \textbf{17.98} & \textbf{12.66} \\
    \bottomrule
  \end{tabularx}
\end{table}

\subsection{AST}

We use three test sets of Zh2en translation to evaluate the performance of AST. Among them, we used the numerical values of SpeechLlaMA and Qwen2-audio in their paper. We tested the BLEU scores of WhisperV3, SeamlessM4T2, and Qwen-audio on three test sets. We use ASR and AST multitask learning methods (MTL) to improve the final performance of AST as in table ~\ref{tab:ast}.

\begin{table}[htbp]
  \caption{AST results: BLEU scores of different large models}
  \label{tab:ast}
  \centering
  \begin{tabularx}{\textwidth}{c|ccccccc}
    \toprule
      \multirow{2}{*}{Models} & Speech &  {Whisper} & {Qwen} & {Qwen2} & {Seamless} & {MooER} & {MooER} \\
       & LLAMA & V3 & audio & audio & M4Y2 & 5k & 5k-MTL \\
    \midrule
    Covost1\_Zh2en & - & 13.5 & 13.5 & - & 25.3 & - & \textbf{30.2} \\
    Covost2\_Zh2en & 12.3 & 12.2 & 15.7 & 24.4 & 22.2 & 23.4 & \textbf{25.2} \\
    CCMT2019\_dev & - & 15.9 & 12.0 & - & 14.8 & - & \textbf{19.6} \\
    \bottomrule
  \end{tabularx}
\end{table}

\section{Discussion}

\subsection{Selection of Encoder}
We selected different encoders on our in house validation test set. We found that if Semi-Supervised Learning (SSL) encoder is used, the encoder needs to participate in training, otherwise the loss will be difficult to converge. Considering the performance, parameter size, and efficiency, we ultimately chose Paraformer as our encoder. In this experiment, we fixed LLM and only the adapter participated in training. (In addition to W2v-Bert2.0 being an encoder, the encoder also participates in training). The results are shown in table ~\ref{tab:enc}.

\begin{table}[htbp]
  \caption{CER of different encoder selected on in house validation dataset}
  \label{tab:enc}
  \centering
  \begin{tabular}{c|ccc}
    \toprule
      & W2v-Bert2 & WhisperV3 & Paraformer \\
      \midrule
      In-house-dev & 11.04\% & 7.20\% & 5.56\% \\
    \bottomrule
  \end{tabular}
\end{table}

\subsection{Granularity of audio modeling}
We attempted to model the granularity of 240ms, 180ms, and 120ms, and found that this parameter is crucial for the fusion of audio and text. In the end, we chose to output an audio embedding every 120ms. In this experiment, we fixed LLM and only the adapter participated in training as shown in table~\ref{tab:gra}.

\begin{table}[htbp]
  \caption{CER of different audio modeling strategies on in house validation dataset}
  \label{tab:gra}
  \centering
  \begin{tabular}{c|ccc}
    \toprule
      & 240ms & 180ms & 120ms \\
      \midrule
      In-house-dev & no-converge & 5.56\% & 5.22\% \\
    \bottomrule
  \end{tabular}
\end{table}

\subsection{Quickly adapt to the vertical domain}

\begin{table}[h!tbp]
  \caption{CER of english test set recognition results}
  \label{tab:adpt}
  \centering
  \begin{tabular}{c|cc}
    \toprule
      ASR-ENG & Paraformer-large & MooER-5k \\
      \midrule
      Librispeech-test-clean & 14.15 & 7.78 \\
      Librispeech-test-other & 22.99 & 15.25 \\
      fleurs\_eng\_dev & 24.93 & 18.89 \\
      fleurs\_eng\_test &  26.81 & 20.41 \\
      gigaspeech\_dev & 24.23 & 23.46 \\
      gigaspeech\_test &  23.07 & 22.09 \\
    \bottomrule
  \end{tabular}s
\end{table}

We train based on a Paraformer large encoder. We used approximately $140\sim 150$ hours of English data and achieved better results on the English test set as in table~\ref{tab:adpt}. At the same time, we attempted to migrate to other tasks, such as AST, and achieved a BLEU score of 25.2 on the Covost2 Zh2en translation test set. We believe that such an approach can also be applied to other low-resource audio understanding task domains, such as minority languages, dialects, etc as shown in table~\ref{tab:adp-ast}.

\begin{table}[htbp]
  \caption{BLEU of different large models on Covost2 Zh2en dataset}
  \label{tab:adp-ast}
  \centering
  \begin{tabular}{c|cccccc}
    \toprule
      \multirow{2}{*}{AST-C2E} & Speech &  {Whisper} & {Qwen} & {Qwen2} & {Seamless} & {MooER} \\
       & LLAMA & V3 & audio & audio & M4T2 & 5k-MTL \\
      \midrule
      Covost2\_Zh2en & 12.3 & 12.2 & 15.7 & 24.4 & 22.2 & 25.2 \\
    \bottomrule
  \end{tabular}
\end{table}

\subsection{Fully utilize the capabilities of large models}

We found that incorporating LLM into audio understanding training can lead to faster and more stable convergence and ultimately achieve better results. Moreover, the final effect will increase as the LLM effect improves as in table~\ref{tab:uti}.

\begin{table}[h!tbp]
  \caption{CER of different LLMs combined with different training strategies}
  \label{tab:uti}
  \centering
  \begin{tabular}{c|cccc}
    \toprule
      \multirow{2}{*}{ASR} & MooER-Qwen1.5 &  MooER-Qwen2 & MooER-Qwen1.5 & MooER-Qwen2 \\
       & frozen & frozen & lora & lora \\
      \midrule
      In-house-dev CER & 5.22\% & 5.17\% & 4.72\% & 4.32\% \\
    \bottomrule
  \end{tabular}
\end{table}

\subsection{Acceleration method}

We optimized the dataloader section, which can increase training speed by 4-5 times under the same configuration. At the same time, we optimized Deepspeed's training strategy based on 5000h of training and reused it in our 8wh internal data training. For training that requires unfrozening the encoder, we use gradient checkpoint to reduce the memory usage. We use the KUAE platform based on Moore Threads for accelerated training of large models.

\section{Demo}

Our demo is built on the domestically produced S4000 GPUs by Moore Threads.
\url{https://mooer-speech.mthreads.com:10077/}

\section{Models}
Github: \url{https://github.com/MooreThreads/MooER}

ModelScope: \url{https://modelscope.cn/models/MooreThreadsSpeech/MooER-MTL-5K}

Huggingface: \url{https://huggingface.co/mtspeech/MooER-MTL-5K}

\begin{ack}
SLAM-LLM, FunASR, Qwen
\end{ack}

\section*{References}

\small

[1] Wang C, Wu A, Gu J, et al. CoVoST 2 and Massively Multilingual Speech Translation[C]//Interspeech. 2021: 2247-2251.

[2] Tang C, Yu W, Sun G, et al. Salmonn: Towards generic hearing abilities for large language models[J]. arXiv preprint arXiv:2310.13289, 2023.

[3] Radford A, Kim J W, Xu T, et al. Robust speech recognition via large-scale weak supervision[C]//International conference on machine learning. PMLR, 2023: 28492-28518.

[4] Barrault L, Chung Y A, Meglioli M C, et al. Seamless: Multilingual Expressive and Streaming Speech Translation[J]. arXiv preprint arXiv:2312.05187, 2023.

[5] Chu Y, Xu J, Zhou X, et al. Qwen-audio: Advancing universal audio understanding via unified large-scale audio-language models[J]. arXiv preprint arXiv:2311.07919, 2023.

[6] SpeechTeam T. FunAudioLLM: Voice Understanding and Generation Foundation Models for Natural Interaction Between Humans and LLMs[J]. arXiv preprint arXiv:2407.04051, 2024.

[7] Wu J, Gaur Y, Chen Z, et al. On decoder-only architecture for speech-to-text and large language model integration[C]//2023 IEEE Automatic Speech Recognition and Understanding Workshop (ASRU). IEEE, 2023: 1-8.

[8] Yang A, Yang B, Hui B, et al. Qwen2 technical report[J]. arXiv preprint arXiv:2407.10671, 2024.

[9] Gao Z, Li Z, Wang J, et al. Funasr: A fundamental end-to-end speech recognition toolkit[J]. arXiv preprint arXiv:2305.11013, 2023.

[10] Hu E J, Shen Y, Wallis P, et al. Lora: Low-rank adaptation of large language models[J]. arXiv preprint arXiv:2106.09685, 2021.

[11] Rasley J, Rajbhandari S, Ruwase O, et al. Deepspeed: System optimizations enable training deep learning models with over 100 billion parameters[C]//Proceedings of the 26th ACM SIGKDD International Conference on Knowledge Discovery \& Data Mining. 2020: 3505-3506.

[12] Ma Z, Yang G, Yang Y, et al. An Embarrassingly Simple Approach for LLM with Strong ASR Capacity[J]. arXiv preprint arXiv:2402.08846, 2024.

[13] Barrault L, Chung Y A, Meglioli M C, et al. Seamless: Multilingual Expressive and Streaming Speech Translation[J]. arXiv preprint arXiv:2312.05187, 2023.

[14] Du J, Na X, Liu X, et al. Aishell-2: Transforming mandarin asr research into industrial scale[J]. arXiv preprint arXiv:1808.10583, 2018.

[15] Panayotov V, Chen G, Povey D, et al. Librispeech: an asr corpus based on public domain audio books[C]//2015 IEEE international conference on acoustics, speech and signal processing (ICASSP). IEEE, 2015: 5206-5210.

[16] \url{https://github.com/wenet-e2e/wenet/tree/main/examples/multi\_cn/s0}

[17] Zhang B, Lv H, Guo P, et al. Wenetspeech: A 10000+ hours multi-domain mandarin corpus for speech recognition[C]//ICASSP 2022-2022 IEEE International Conference on Acoustics, Speech and Signal Processing (ICASSP). IEEE, 2022: 6182-6186.

[18] Conneau A, Ma M, Khanuja S, et al. Fleurs: Few-shot learning evaluation of universal representations of speech[C]//2022 IEEE Spoken Language Technology Workshop (SLT). IEEE, 2023: 798-805.


\appendix

\section{LORA Configure}

\begin{lstlisting}[language=Python, caption=Lora Configure]
lora_r = 64
lora_alpha = 16
target_modules = [
        "q_proj",
        "k_proj",
        "v_proj",
        "o_proj",
        "up_pro
]
\end{lstlisting}

\section{DeepSpeed Configure}
\begin{lstlisting}[language=Python, caption=Lora Configure]
"train_micro_batch_size_per_gpu": 8,
"gradient_accumulation_steps": 2,
"optimizer": {
    "type": "Adam",
    "params": {
        "lr": 1e-4
    }
},
"scheduler": {
    "type": "WarmupDecayLR",
    "params": {
        "total_num_steps": 1000000,
        "warmup_max_lr": 0.0001,
        "warmup_num_steps": 1000
    }
},
"bf16": {
        "enabled": true
    },
"zero_optimization": {
    "stage": 2,
    "allgather_partitions": true,
    "allgather_bucket_size": 2e8,
    "overlap_comm": true,
    "reduce_scatter": true,
    "reduce_bucket_size": 2e8,
    "contiguous_gradients": true
}
\end{lstlisting}

\end{document}